\documentclass[11pt,twocolumn]{article}
\usepackage{graphicx}
\usepackage{amsmath}   % From the American Mathematical Society
\usepackage{url}
\usepackage{multirow}
\usepackage[colorlinks=true,linkcolor={black},urlcolor={black}]{hyperref}
\usepackage{ftnright}
\newcommand{\be}{\begin{equation}}
\newcommand{\ee}{\end{equation}}
\begin{document}

\title{Impact of Cognitive Radio on Future Management of Spectrum\\
(Invited Paper)}

\date{\today}
\author{Maziar Nekovee \\ 
BT Research, Polaris 134, Adastral Park,\\
Martlesham, Suffolk, IP5 3RE, UK\\
and
\\
Centre for Computational Science, University College London\\
20 Gordon Street, London WC1H 0AJ, UK \\ 
maziar.nekovee@bt.com}
\maketitle

\begin{abstract}
Cognitive radio is a breakthrough technology which is expected  to have a profound impact
on the way radio spectrum will be accessed, managed and shared in the future.
In this paper I examine some of the implications of 
cognitive radio for future management of spectrum. Both a near-term 
view involving the opportunistic spectrum access model 
and a longer-term view involving a self-regulating 
dynamic spectrum access model within a society of cognitive radios are discussed.
\end{abstract}

\section{Introduction}
Spectrum availability in current wireless communication systems is
decided by regulatory and licensing bodies. The mainstream spectrum management approach,
adopted around the world, is based on a static spectrum
allocation model known as Command \& Control. In this model 
the available radio spectrum is divided into fixed and 
non-overlapping blocks, separated by so-called guard bands, assigned to different services and wireless technologies. These blocks are then licensed for 
exclusive use to carriers, radio and TV broadcasters, specialised mobile radio
operators, the military and public safety applications. Equivalent Isotropically Radiated Power (EIRP) and out-of-band emission 
(i.e interference to neighbouring frequencies) within each band  are rigidly defined 
in the licences' terms. Regulation 
takes care of protection against interference and provides  a limited support 
for coexistence capabilities.

Static spectrum assignment combined with the phenomenal
growth in demand for spectrum are the  reasons behind the commonly
shared feeling of spectrum scarcity.  
Cognitive radio (CR) 
\cite{mitola1,mitola-book,fette}
is currently considered as one of the most promising solutions to the
aforementioned scarcity problem by enabling a highly dynamic,
device-centric spectrum access in future wireless communication
systems. A CR can adapt the operation parameters of its radio
(frequency band, modulation, coding etc) and its transmission or
reception parameters on the fly based on cognitive interaction with
the wireless environment in which it operates. 

Advances brought about by cognitive radio technology and
software-defined radio (SDR) not only hold great promise for a much more efficient dynamic 
access to spectrum, they are also driving and enabling  radically 
new models of spectrum management.  In the US FCC (Federal Communications 
Commission) proposed to allow
opportunistic access to TV bands by cognitive radios already in 2004
\cite{fcc-cr}. Prototype cognitive radios operating in this 
mode have been put forward by Philips, Microsoft \cite{cr-ms}
and Motorola. Furthermore, 
the emerging 802.22 standard for cognitive radio access to TV bands
\cite{ran1,ran2} is at its final stage of development. Adopting a
more cautious approach, the UK regulator, Ofcom (Office of
Communications), initially 
refrained from allowing licensed-exempt operations of cognitive radio \cite{ofcom-sdr}. 
However, in what is potentially a radical shift in policy, 
in its recently released Digital Dividend Review Statement 
\cite{ofcom-ddr} Ofcom is proposing to ``allow licence exempt use of interleaved spectrum for cognitive
devices.'' \cite{ofcom-ddr}. Furthermore Ofcom has
``decided not to set aside any of the digital dividend exclusively for
licence-exempt use'', arguing that  ``the opportunity cost of setting
aside spectrum just for licence-exempt use would be high, and the
additional benefits would be limited given the prospects of cognitive
access to interleaved spectrum'' \cite{ofcom-ddr}.
 
With both the US and the UK adapting the opportunistic spectrum access
(OSA) model we can expect that, if successful, this new paradigm will become mainstream in the toolbox
of spectrum regulators worldwide. However, future implications of
cognitive radio may go well beyond the adaptation of the 
OSA model. In the long run 
the advanced capabilities of cognitive radios may allow the Command \&
Control spectrum management model to be entirely replaced  by a
radically new dynamic spectrum access model within a society
of cognitive devices. In this model cognitive and highly 
reconfigurable wireless devices self-regulate their access and sharing
of spectrum on behalf of the users, based on a minimum set of imposed policies
and  through a continuous process of communications, negotiation, trading and  cooperation.

The aim of this paper is to  examine in some detail such
future modes of spectrum management which are both driven and enabled 
by the emergence of cognitive radio and technologies  such
as software-defined radio \cite{fette} and spectrum pooling \cite{specu2}.

The rest of this paper is organised as follows.
Since the concept of cognitive radio means different things to different people, we provide in section II a 
definition of CR which will be used through the rest of the paper.
Section III describes and critically explores opportunistic access
with cognitive radios, with particular emphasise given to cognitive
operation in UHF/VHF TV bands. In section IV we examine the paradigm of 
dynamic spectrum access with cognitive radio, and put forward the notion of a self-regulatory  
access model, which shows interesting analogies with the way 
pedestrian traffic is self-regulated in our societies. 
We conclude  this  paper in Section V with conclusions.

\section{Defining Cognitive Radio}
The term cognitive radio (CR) was first introduced by Mitola \cite{mitola1} as 
``the point in which wireless personal digital assistants (PDAs) and
the related networks are sufficiently computationally intelligent
about radio resources and related computer-to-computer communication
to: (a) detect user communications needs as a function of user context
and (b) to provide radio resources and wireless services most
appropriate to those needs''. 
Recently the term cognitive radio has been used in a narrower sense for radio systems that have adaptive spectrum awareness. The FCC, for example, defines it in the following 
way \cite{fcc-cr} (a similar definition is also used by Ofcom):
\\
\begin{quote}
``A Cognitive Radio (CR) is a radio that can change its transmitter parameters based on interaction with the environment in which it operates. The majority of cognitive radios will probably be SDR (Software Defined Radio) but neither having software nor being field programmable are requirements of a cognitive radio.''
\end{quote}
From the above it can be seen that there is no unique definition for cognitive radio and depending on the focus (e.g. users requirements versus system requirements) and applications different definitions can be put forward. 
The two main characteristics that come forward in most definitions, however, 
are reconfigurablity and intelligent adaptive behaviour.
   
In the rest of this paper we shall adopt the following definition:
\begin{quote}
{\it
A cognitive radio is an autonomous radio that can intelligently adapt its operational characteristics (frequency, waveform, modulation, power, antenna) on the fly, in response to changes to its  electromagnetic environment 
while complying with  spectrum policies, with the aim of optimally  meeting  user's  requirements for 
wireless access.}
\end{quote}
In the above by intelligent adaptive behaviour we mean the ability to adapt without being a priori programmed to do this, i.e. via some form of learning. For example, a handset that learns a radio frequency map in its surrounding could create a location-indexed RSSI vector (Latitude, Longitude, Time, RF, RSSI) and uses a machine-learning algorithm based on which it switches its frequency band or base station as the user moves \cite{mitola-book}. 
Furthermore, the term 'radio' is taken to mean any system that communicates with other systems via a modulated 
signal within the radio frequency spectrum.
 
From the above definition it follows that cognitive radio functionality requires the following capabilities:
\begin{itemize}
\item 
{\bf Flexibility and agility}: the ability to change the waveform and other radio operational parameters 
on the fly.\\ 
This is to a limited extent possible with the current multi-frequency multi-access radios. However, full flexibility  become possible when cognitive radios are built on top a software-defined radio. An SDR is a radio in which the properties of carrier frequency, signal bandwidth, modulation and network access are defined by software. In addition to SDR, another important requirement to achieve flexibility, which is often overlooked, is reconfigurable and/or 
wideband antenna technologies to support wide-band spectrum agility.

\item
{\bf Sensing}: the ability to observe and measure the state of the environment, including spectral occupancy. 
\\
Sensing is necessary if the device is to change its operation based on its current knowledge of Radio Frequency (RF) environment.
\item
{\bf Learning and adaptability}: the ability to analyse sensory input , recognise patterns and modify internal operational behaviour based on the resultant analysis of the new situation,  
not only based on pre-coded algorithms and heuristics but also as a result of a learning mechanism. 
\\
The IEEE 802.11 MAC layer allows a device to adapt its transmission activity to channel availability that it senses. However, this is achieved using a pre-defined listen-before-talk and exponential backoff algorithm, and so an 802.11 device is not cognitive. 
\end{itemize}
In addition to the above core abilities the operation of a cognitive radio often requires 
{\it location awareness} in order to be able to respond to spatially variant regulatory policies or spatially variant spectrum availabilities.    

\section{Opportunistic spectrum access with cognitive radio}
Cognitive radio technology  is being intensively researched as a key
enabler for the opportunistic spectrum access 
model \cite{horn,maz1-cr}.
In this operational mode a
cognitive radio acts as a spectrum scavenger. It continuously performs
spectrum sensing over a 
range of frequency bands, dynamically identifies unused {\it licensed}
spectrum (the so-called White Spaces), and then operates in this
spectrum at times and/or locations  when/where it is not used by incumbent
radio systems. In this mode a cognitive radio may coexist with the primary users
either on a not-to-interfere basis or on an easement basis, which
allows secondary transmissions as long as they are below some
acceptable interference threshold.  We note that opportunistic
spectrum access can take place both on a temporal and a spatial
basis. In temporal opportunistic access a cognitive radio
monitors the activity of the licensee in a given location and uses the
licensed frequency at times that is idle. An example of this is the
operation of cognitive radio in the UMTS bands \cite{umts}.
In spatial opportunistic access 
low-power cognitive devices identify geographical regions  
where certain licensed bands are unused and access these bands without causing harmful
interference to the operation of the incumbent in nearby regions. The
operation of cognitive radios in TV bands, for example, is
primarily based on spatial opportunistic access to these bands, and
will be discussed further in the following sections. 

\subsection{Cognitive access to TV bands}
Broadcast television services operate in licensed channels in the VHF and UHF portions of the radio spectrum. The regulatory 
rules in most countries prohibit the use of unlicensed devices in TV
bands, with the exception of remote control, medical telemetry devices
and cordless microphones. In the US and the UK regulators are currently in the
process of requiring TV stations to convert from analogue to digital
transmission. This {\it Digital Switchover} is  expected to
be completed in the US in 2009 and in the UK in 2012. A similar
switchover process  is also underway or being planned in the rest of
the EU and many other countries around the world.

After Digital Switchover a portion  of TV analogue channels
become  entirely vacant  due to the higher spectrum efficiency of digital TV (DTV). These
cleared channels will then be reallocated by regulators to other services,
for example through auctions. In addition, after the DTV
transition there will be typically a number of TV channels in a  given geographic area that are not being used by DTV
stations, because such stations would not be able to operate without
causing interference to co-channel or adjacent channel stations. These
requirements are based on the assumption that stations operate
at maximum power. However, a transmitter operating on a vacant TV
channel at a much lower power level would not need a great separation
from co-channel and adjacent channel TV stations to avoid causing
interference. Low power unlicensed  devices can operate on vacant channels in
locations that could not be used by TV stations due to interference
concerns \cite{new-america}.
These vacant TV channels are known as TV White Spaces, or
interleaved spectrum in the parlance of the UK regulator. Fig. 1 shows, as an example,
the chart of the UK's analogue TV frequency bands and how these will be
divided after digital switchover into cleared and interleaved spectrum \cite{ofcom-ddr}.

The FCC adopted a Notice for Rulemaking (NRM) in 2004 proposing to allow unlicensed radio transmitters to operate in the broadcast TV spectrum at locations where that spectrum is not being used \cite{fcc-cr}. The proposed new rules would, in principle allow the operation of both fixed and portable broadband devices on a 
non-interference basis. Very recently, a similar statement proposing to allow 
unlicensed operation of  cognitive radio devices in interleaved
spectrum was released by Ofcom as part of its Digital Dividend
Review (DDR) \cite{ofcom-ddr}.  

Opportunistic operation of  cognitive
radios in TV bands, however, is conditioned on the ability of these
devices to avoid harmful interference to licensed users of these
bands, which in addition to DTV include also wireless microphones
\cite{ new-america}.
FCC discusses three methods for ensuring that unlicensed TV band devices  do not cause harmful interference to  incumbent: control signals, position determination, and cognitive radio.  In the following we shall briefly discuss these methods. 

\begin{figure}
\centering
\begin{tabular}{@{}ccl@{}}
\includegraphics[width=3.5in]{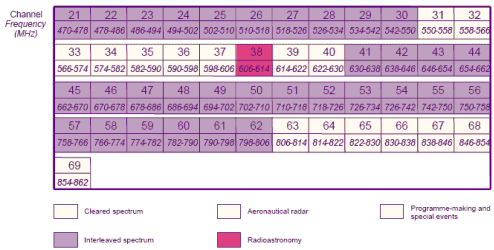} 
\end{tabular}
\caption{The available UHF TV spectrum in the UK after digital switchover, 
showing both the interleaved and cleared channels \cite{ofcom-ddr}  }
\label{fig:specpie}
\end{figure}

With the control signal method, unlicensed devices only transmit if
they receive a control signal (beacon) identifying vacant channels
within their service areas. The signal can be received from a TV
station, FM broadcast station, or TV band fixed unlicensed
transmitter. Without reception of this control signal, no
transmissions are permitted. In the position determination method, an
unlicensed device incorporates a GPS receiver to determine its
location and access a database to determine the TV channels that are
vacant at that location.  There are at least two issues associated with this
method \cite{new-america}: 1) the accuracy and completeness of the database and 2) the ability of  unlicensed device to determine ts location.   Finally, in the cognitive radio method, unlicensed devices autonomously detect the presence of TV signals and only used the channels that are not used by TV broadcaster.   
	  
\subsection{How much spectrum are we talking about?}
The fundamental reason why TV spectrum has attracted much interest is
an exceptionally attractive combination of bandwidth and
coverage. Signals in TV bands, travel much further than both the WiFi and 
3G signals and penetrate buildings more readily. This in turn means
that these bands can be used for a very wide range of potential new
services, including last mile wireless broadband in urban environments, broadband wireless
access in rural areas \cite{ran1,ran2}, new types of mobile
broadband and wireless networks for digital homes. 
On the other hand, the available spectrum for cognitive access varies 
from location to location and depends strongly on the population density. 
In rural areas, where the population density is low, there is also  low concentration of  TV transmitters, and a large number of TV channels 
could be vacant. On the other hand, in urban areas the density of TV 
transmitters could be very high, and therefore  the number of unused TV channels
(White Spaces) at a given location could be very limited.
In order to to asses the importance  of cognitive access to TV bands it is therefore important to have realistic estimates of the amount of spectrum (spectrum opportunity) that is associated with this mode of access.

Spatial  variations in TV White Spaces can  be assessed from a TV transmission 
database, which specifies the location of transmitters, the frequencies 
they use at a given location, and the typical coverage area of the
transmitter. Fig. 2 shows schematically a typical setup for the operation of 
a cognitive radio base station which operates in a given location 
in TV White Spaces which are available at that location.
The CR transmission  should not cause harmful interference to TV receivers
both within the coverage area of nearby transmitters, and at the edge 
of this area. To achieve this the CR device can
transmit on the TV bands used by these transmitters only if its position
is a  minimum ``keep-out'' distance away from the  edge of
their coverage area \cite{ran2}.

\begin{figure}
\centering
\begin{tabular}{@{}ccl@{}}
\includegraphics[width=3.5in]{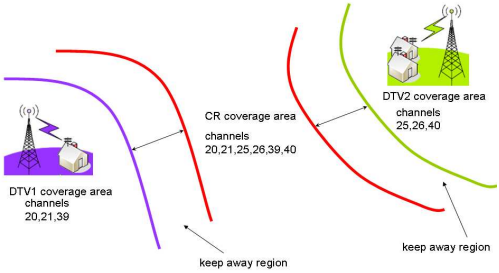} 
\end{tabular}
\caption{Opportunistic access to interleaved TV spectrum (White Spaces) 
by cognitive radios.}
\label{fig:specpie}
\end{figure}
In a simplified picture, based on the pathloss model \cite{pathloss}, the
keep-out distance can be obtained as follows.  Denote with $R_{tv}$ the
maximum coverage radius of the TV station, and with $P_{cr}$ and $P_{tv}$ the
transmit power of the TV transmitter and the CR transmitter,
respectively.  Then, in order to avoid interference with TV receivers that are 
at  the edge of the coverage area, we must have: 

\begin{equation}
\frac{ P_{tv} /R_{tv}^\alpha }
      {P_{cr}/R_{cr}^\alpha }\geq \beta_{th},
\end{equation}
where $\beta_{th}$ is the sensitivity threshold of a TV
receiver, and $\alpha$ is the pathloss exponent. This yields: 
\begin{equation}
R_{cr}\geq \left 
( \beta_{th} \frac{P_{cr}}{P_{tv}}\right)^{1/\alpha} R_{tv}.
\end{equation}
Consequently, a CR device at location ${\bf r}$ can use the frequencies
associated with a TV station located at ${\bf R}_j$ only if 
\begin{equation}
|{\bf r}-{\bf R}_j| \geq  \left[ 1+
\left ( \beta_{th} \frac{P_{cr}}{P_{tv}}\right)^{1/\alpha}\right] R_{tv}.  
\end{equation}

Repeating the above procedure for every TV transmitter, 
one can obtain the total number of TV transmitters
in a given region whose associated frequencies can be used by a CR
operating with a specified transmit power $P_{cr}$ at location ${\bf r}$, 
from which the total number of TV
bands available for opportunistic access can be obtained from information 
of TV transmitters' frequency allocation plans\footnote
{We note that usually one requires that channels adjacent to an occupied 
TV channel should also be avoided  by CR  devices  in order 
to eliminate adjacent channel interference effects \cite{tv-alex}.}. 

Using a similar approach, but taking into account the actual service contours
of TV transmitters, Brown and
Sicker \cite{Siker} have made  estimates of the spectrum available to CR
devices at a number of locations in the US,  as a function of the
transmit power  (or equivalently transmission range) of a cognitive
radio. Using New York City, as an example of a highly populated urban area and 
Buffalo, as an example of a rural area, they show that the availability 
of spectrum for cognitive access   greatly depends on location and transmission range.
In the case of Buffalo, about $50$ TV channels ($300$ MHz) would be available
for a CR transmitting to distance of up to $10$ km. On the other hand, 
in New York, the maximum number of available channels is only $4$ ($24$ MHz) 
and drops sharply beyond a $10$ km transmission range.  

We are currently investigating  spatial 
variations in  the  TV  White Spaces (interleaved spectrum) which will
become available in the UK. A preliminary study by Ofcom, however,
indicates that 
``at any one location, around $100$ MHz on average is not being used by DTT (Digital Terrestrial Television)
and could, in principle be used by licence-exempt devices'' \cite{ofcom-ddr}. 
Comparing this amount of spectrum with, e.g., the total 
UK 3G spectrum ($\sim 75$ MHz) it can be seen that the potential spectrum 
opportunity available for cognitive access is significant.
Effective utilisation of this opportunity for commercial applications, 
however, requires further research in the development of new
algorithms for incumbent detection,  as well as new
approaches for effective pooling of discontiguous TV channels \cite{ofdm}.

\section{Next generation Dynamic Spectrum Access with Cognitive radio}

\begin{figure}
\centering
\begin{tabular}{@{}ccl@{}}
\includegraphics[width=2.5in]{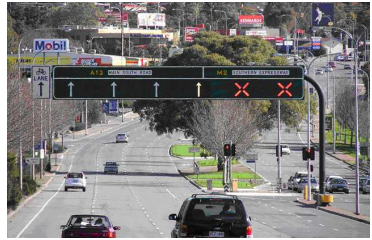} \\ 
\includegraphics[width=2.5in]{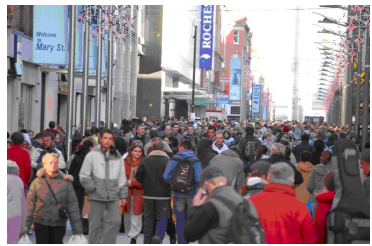} 
\end{tabular}
\caption{The current spectrum management regime is analogous to 
how road traffic is currently controlled (left panel), with clearly defined 
lanes and centrally controlled signals. The emergence of cognitive 
radio may result in a much more distributed management model analogous
to the  way pedestrian traffic is self-regulated.}
\label{fig:specpie}
\end{figure}
As spectrum liberalisation \cite{ofcom-sur}, trading \cite{cave} 
and a technology-based approach to interference management becomes increasingly
mainstream, the radio technologies that allow the paradigm of 
Dynamic Spectrum Access (DSA) will enable more dynamic and efficient management 
of the spectrum resources \cite{maz-dsa}.
The IEEE P1900 working group defines DSA
in the following way \cite{p1900}
\\
\begin{quote}
``Technique by which a radio (system) dynamically adapts to select operating spectrum to use available (in local time-frequency space) spectrum holes with limited spectrum use rights.''
\end{quote}
The concept of DSA is not new. Indeed, BT's Fusion phone and the
Unique phone from Orange are two examples of consumer products which
dynamically select spectrum based on the operating environment. Both
products employ a technology known as Unlicensed Mobile Access (UMA)
and are strictly limited to a choice between either WiFi/Bluetooth or
GSM/3G bands. Although this choice happens on a dynamic basis, the
spectrum landscape itself remains static. Dynamic spectrum access
techniques need not be limited to simple choices between alternative
bands.

The introduction of spectrum liberalisation and trading \cite{cave,sverrir}, 
combined with increased ability of devices 
to change their radio operations on the fly, may lead to far more
adventurous forms of DSA \cite{lehr-chapin,maz-dsa}. Devices will be
able to dynamically vary their operating spectrum over a wide range of
frequency bands, allowing them to access these
bands on a just-in-time basis and according to their market price. This may happen either upon instruction from a base station which 
dynamically negotiate and acquire spectrum  on behalf of user devices,
or autonomously  by devices themselves \cite{maz1-cr}.
Some, less congested, bands may be available at a significantly lower cost (or even free) than the more popular bands and devices would be able to utilise these, ensuring that users remains connected at a price that supports their utility. 

When combined with higher levels of built-in cognitive intelligence,
advanced  pooling techniques and reconfigurable antenna technologies
the concept of DSA can be taken still further. 
The entire available spectrum may be  divided
into a very large number of equally-sized  {\it elementary subchannels} (ESC).
Depending on  their requirements devices may  
pool together and  utilise  a  number of (not necessarily contiguous)
ESCs, and then  vacate some or  all of these  when they are not longer
required or when other more suitable ones  become available.
The static and highly fragmented  spectrum landscape of today
will be replaced with a highly dynamic quasi-continuum landscape,
in which  each device navigates its  own way  using  its
cognitive and adaptive capabilities, in order to
optimally meet user's requirement for wireless access.

To illustrate the above concept, we can draw an analogy with vehicular and pedestrian traffic.
The recently emerging DSA systems can be compared to how vehicular
traffic on a multi-lane highway operates. Here 
vehicles  are free to choose which lane they occupy and can
change lane, but are restricted in their manoeuvres to switching 
between a  number of marked lanes.  
In contrast, future DSA systems  are more akin to the pedestrian motion in a busy pedestrianised street. 
Here there are no well defined lanes and only a  minimum set of rules are imposed externally.
Each pedestrian autonomously navigate and manoeuvre his/her own way in the crowd, using a cognitive cycle which 
involves choosing a position and direction of motion that best satisfies his/her own personal goals and 
dynamically adjusting his/her movements and pace based on his/her perception/prediction of the 
movement of other pedestrians and the restrictions imposed by the
environment.

\section{Conclusions}
In this paper we examined how the emergence of cognitive radio may impact the way spectrum will be managed and regulated 
in both the near (next $5$ years) and more distant (next $10-20$
years) future. 
We showed that the emerging paradigm of opportunistic spectrum access
by cognitive radios is already 
making an important  impact on the way the highly valuable spectrum in
the VHF/UHF TV bands will be accessed, managed and shared 
in the US and the UK. 
If successful, this new mode of access may be adopted by  other regulators
around the world. It may  also be extended to other portions of the licensed
spectrum, such as  the 3G and the newly released WiMAX bands.
opportunistic access to such bands, however, may need to be supported
by  new mechanisms such as real-time trading and spectrum leasing
which create  economic incentive for the incumbents to allow
secondary utilisation of their spectrum when it is idle. We saw that
Opportunistic access by cognitive radios is also threatening the current 
licenced-exempt model where non-cognitive device operate in free-for-all bands which are set aside for their
operations.   

In the long run, with rapid advancement of  device
reconfigurability and cognitive capabilities, we  may see the spectrum
management model based on rigidly defined frequency
bands disappear altogether.  Instead, the spectrum will be managed as  a
continuum-type resource whose access and sharing is largely
self-regulated by a society of cognitive devices which are engaged, on behalf of
their users, in  a continuous process of  communications, negotiations, trading and
cooperation.

\section*{Acknowledgements}
The views expressed in this paper are those of the author and do not necessarily reflect those of BT.

\end{document}